\documentclass{ngsm2024} 

\title[An All-MLP Sequence Modeling Architecture That Excels at Copying]{An All-MLP Sequence Modeling Architecture That Excels at Copying}

\optauthor{%
\Name{Chenwei Cui}\textsuperscript{1}, 
\Name{Zehao Yan}\textsuperscript{2}, 
\Name{Gedeon Muhawenayo}\textsuperscript{1}, 
\Name{Hannah Kerner}\textsuperscript{1}\thanks{Corresponding author: hkerner@asu.edu; Code: github.com/FrankCuiCN/causal-relation-networks}\\
\addr \textsuperscript{1}Arizona State University; \textsuperscript{2}Colorado State University}

\begin{document}

\maketitle

\begin{abstract}
Recent work demonstrated Transformers' ability to efficiently copy strings of exponential sizes, \linebreak distinguishing them from other architectures.
We present the Causal Relation Network (CausalRN), \linebreak an all-MLP sequence modeling architecture that can match Transformers on the copying task.\linebreak
Extending Relation Networks (RNs), we implemented key innovations to support autoregressive sequence modeling while maintaining computational feasibility.
We discovered that exponentially-activated RNs are reducible to linear time complexity, and pre-activation normalization induces an infinitely growing memory pool, similar to a KV cache.
In ablation study, we found both exponential activation and pre-activation normalization are indispensable for Transformer-level copying.\linebreak
Our findings provide new insights into what actually constitutes strong in-context retrieval.

\end{abstract}


\section{Background}
Recently, \cite{copy2024jelassi} proposed an elegant exam, the copying task, to distinguish Transformers' \cite{transformers2017vaswani} retrieval ability from mainstream alternative architectures such as Recurrent Neural Networks (RNNs) \cite{lstm1997hochreiter, rwkv2023peng} and State-Space Models (SSMs) \cite{s4ssm2022albert, mamba2023gu}. The uncomfortable situation is, on some important tasks, the specific construct of Transformers is the one and only working combination. We therefore ask: \textit{Can we find a non-trivial mechanism that matches Transformers on the copying task?}

\paragraph{Notation}
We use bold lowercase letters for vectors and bold uppercase letters for matrices.
$[\mathbf{x}_i; \mathbf{x}_j]$ represents the vertical concatenation of \(\mathbf{x}_i\) and \(\mathbf{x}_j\). 
We use \(\circ\) for element-wise product.
To describe the size of a neural network, we use $d_e$ for embedding size and $d_h$ for the number of hidden neurons.

\paragraph{Multi-Layer Perceptrons (MLPs)}
Consider a non-linear element-wise activation function $\psi$.
For input
$\mathbf{x} \in \mathbb{R}^{d_{e}}$
and parameters
$\mathbf{b}_{in} \in \mathbb{R}^{d_h},
\mathbf{b}_{out} \in \mathbb{R}^{d_e},
\mathbf{W}_{in} \in \mathbb{R}^{d_h \times d_e},$
and
$\mathbf{W}_{out} \in \mathbb{R}^{d_e \times d_h},$
\textit{Single-Hidden-Layer Multi-Layer Perceptron (MLP)} \cite{neuralnet1957rosenblatt}
is defined as
\begin{equation}
\label{eq_mlp}
f_\theta(\mathbf{x}) =
\mathbf{W}_{out} \psi(\mathbf{W}_{in} \mathbf{x} +
\mathbf{b}_{in}) +
\mathbf{b}_{out}.
\end{equation}
MLPs only accept fixed-dimension input vectors. This limits their use for sequence modeling \cite{deepsets2017zaheer}.

\paragraph{Relation Networks (RNs)}
For inputs
$X = \{\mathbf{x}_1, \mathbf{x}_2, ..., \mathbf{x}_n\}$
$\mathbf{x}_i \in \mathbb{R}^{d_{e}},$ and
$f_\theta:\mathbb{R}^{2d_e} \rightarrow \mathbb{R}^{d_e},$
\textit{Relation Network} \cite{relationnet2017santoro} is defined as
\begin{equation}
\label{eq_relation_network}
\mathbf{y} = \frac{1}{n^2} \sum_{i=1}^{n} \sum_{j=1}^{n} f_\theta([\mathbf{x}_i; \mathbf{x}_j]).
\end{equation}
Relation Networks (RNs) \cite{relationnet2017santoro} explicitly model pairwise dependencies within a set of input vectors.
A major strength of RNs is their simplicity: they only involve MLPs and summation of vectors.
However, RNs cannot perform autoregressive sequence modeling, since they output single vectors.

\begin{figure}[t]
    \centering
    \includegraphics[width=5.2in]{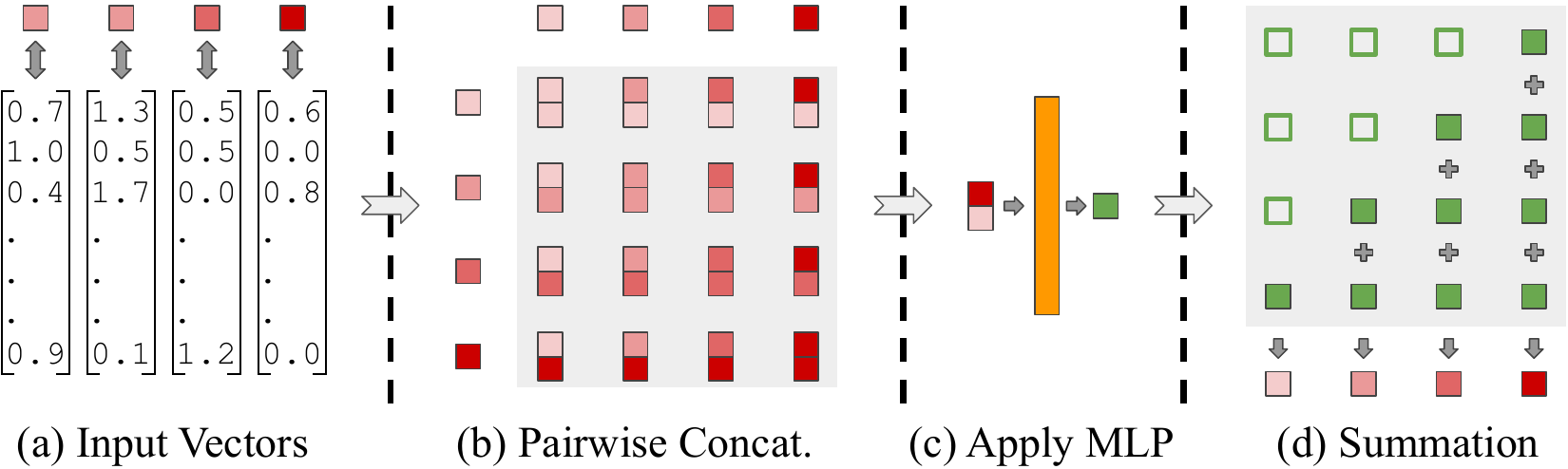}
    \caption{
    Illustration of a CausalRN block. Each square is a feature vector.
    (a) First, we arrange a number of input vectors in sequence.
    (b) Second, the vectors are pairwise concatenated with each other.
    (c) We then apply an MLP to each one of the concatenation.
    (d) Finally, we add the vectors along an axis to obtain the output vectors.
    This step is causally masked.
    }
    \label{figure_causal_rn}
\end{figure}

\section{Causal Relation Networks}
\label{section_causal_relation_networks}

\paragraph{Equivariance and Causality}
Equivariant architectures that are causally-masked allow for efficient autoregressive training \cite{transformers2017vaswani}.
We relax the inner summation of Eq.~\ref{eq_relation_network} to make RNs equivariant.
Causal masking is achieved by capturing only the causal relationships between the feature vectors.
Specifically, for vectors
$X = \{\textbf{x}_1, \textbf{x}_2, ..., \textbf{x}_n\}$
where $\mathbf{x}_i \in \mathbb{R}^{d_{e}},$ and
$f_\theta:\mathbb{R}^{2d_e} \rightarrow \mathbb{R}^{d_e},$
we define \textit{Causal Relation Network (CausalRN)} as
\begin{equation}
\label{eq_causalRN}
\mathbf{y}_{j} = \frac{1}{j} \sum_{i=1}^{j} f_\theta([\mathbf{x}_i ; \mathbf{x}_j]).
\end{equation}
Figure~\ref{figure_causal_rn} offers an illustration. CausalRNs are analogous to Causal Transformers such as GPTs \cite{gpt2018radford}.

\paragraph{Computation Consideration}
Applying a quadratic number of MLPs can be computationally prohibitive.
We notice the input and output layers of the MLP can be pre-computed in \(O(n)\) and reused.
Expanding and rearranging Eq.~\ref{eq_causalRN} and Eq.~\ref{eq_mlp}, splitting \(\mathbf{W}_{in} \in \mathbb{R}^{d_h \times 2d_e}\) into \(\left(\mathbf{W}_{left} ~~ \mathbf{W}_{right}\right)\), we get

\begin{equation}
\label{eq_computation_consideration}
\mathbf{y}_{j} = \mathbf{b}_{out} + \frac{1}{j} \mathbf{W}_{out} \sum_{i=1}^{j}\psi(\mathbf{W}_{left} \mathbf{x}_i + \mathbf{W}_{right} \mathbf{x}_j + \mathbf{b}_{in}).
\end{equation}
Therefore, all affine transformations can either be pre-computed or postponed until after summation.

\paragraph{Exponential Activation}
Based on Eq.~\ref{eq_computation_consideration}, we make a surprising discovery that by using \(\exp(x)\) as the activation function, CausalRNs can be linearized exactly.
To show this, we focus on the summation part. Setting
\(\mathbf{p}_i = \mathbf{W}_{left} \mathbf{x}_i\)
and
\(\mathbf{q}_j = \mathbf{W}_{right} \mathbf{x}_j + \mathbf{b}_{in}\),
by the exponential property,
\begin{equation}
    \sum_{i=1}^{j} \exp(\mathbf{p}_i + \mathbf{q}_j) = \sum_{i=1}^{j} \exp(\mathbf{p}_i) \circ \exp(\mathbf{q}_j) = \exp(\mathbf{q}_j) \circ \sum_{i=1}^{j} \exp(\mathbf{p}_i).
\end{equation}
This means that we can pre-compute and reuse \(\sum_{i=1}^{j} \exp(\mathbf{p}_i)\) for any \(j\), allowing for \(O(n)\) training and \(O(1)\) inference.
We compare \(\exp(x)\) with commonly used activation functions in Section~\ref{section_experiments}.
In Appeidix~\ref{appendix_implement_LRN}, we share code snippets for numerically stable implementation of Linear CausalRNs.
We share further experimental results of our linear models in Appendix~\ref{appendix_crn} and Appendix~\ref{appendix_birn}.

\paragraph{Post-Reduction Normalization}
We propose post-reduction normalization to ensure the vector sum has a stable variance.
This is achieved by applying Layer Normalization \cite{layernorm2016ba} right after the summation step in Eq.~\ref{eq_computation_consideration}.
We test post-reduction normalization and its effect in Section~\ref{subsection_ablation_study}.

\paragraph{Pre-Activation Normalization}
While Linear CausalRNs allow for fast training and inference,
their memory reduce to fixed vectors, similar to RNNs and SSMs. This hurts retrieval ability \cite{rnnnottransf2024wen, copy2024jelassi}.
We propose to use pre-activation normalization to enforce an irreducible memory pool similar to a KV cache \cite{transformers2017vaswani}.
This is possible because \(\exp{(\mu(x+y))}\), where \(\mu\) represents normalization, is no longer decomposable.
We test pre-activation normalization in Section~\ref{subsection_ablation_study}.
We also test an approximate version, \(\exp{(\mu(x)+\mu(y))}\), which preserves decomposability.

\section{Experiments}
\label{section_experiments}
\paragraph{Setup} We construct the copying task following \cite{copy2024jelassi}. We sample an alphabetical random string and require models to repeat it.
There are three special tokens: \texttt{<BOS>} to signal the beginning of a random string, \texttt{<SEP>} to signal the end of the random string, and \texttt{<EOS>} to signal the end of the decoding process.
The string lengths range from 16 to 256, with the corresponding context window ranging from 34 to 514.
For evaluation, we calculate the cross-entropy loss and average accuracy from 320 online samples per iteration.
We compare CausalRNs with Linear CausalRNs, Transformers, and Linear Transformers.
In Appendix~\ref{appendix_implementation_details}, we report further implementation details.

\paragraph{Ablation Study}
\label{subsection_ablation_study}
We performed a careful ablation study to evaluate the contribution of our proposed components. All experiments are performed on strings of size 128.

\begin{figure}[h]
    \centering
    \includegraphics[width=6in]{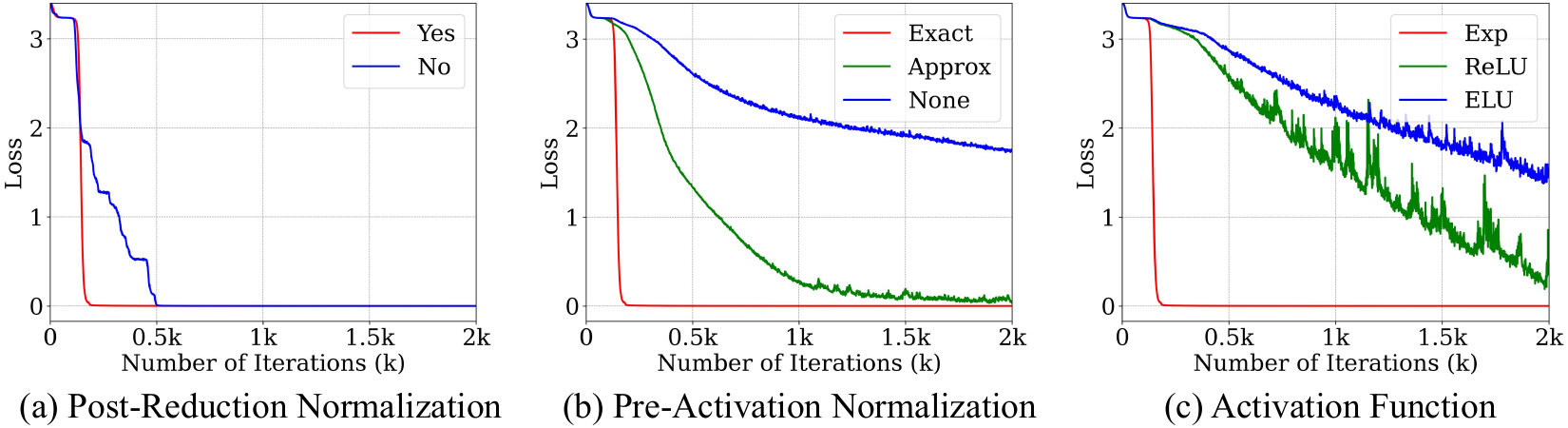}
    \caption{
    Ablation study results.
    The red lines correspond to our proposed CausalRN configuration.
    The blue or green lines indicate the removal or changing of one of the components.
    }
    \label{figure_ablation}
\end{figure}

In Figure.~\ref{figure_ablation}~(a), we see that post-reduction normalization improves the training stability. This verifies our assumption that normalization after vector summation stabilizes CausalRNs.

In Figure.~\ref{figure_ablation}~(b), we observe that only exact pre-activation normalization induces an obvious phase change phenomenon \cite{grokking2023nanda} near the 200\textsuperscript{th} iteration. This highlights the importance for sequence modeling architectures to maintain non-reducible memory pools, rather than fixed memory vectors.

In Figure.~\ref{figure_ablation}~(c), we see that the use of the exponential activation function fundamentally accelerates convergence, while both ReLU and ELU show gradual and linear descent.
This phenomenon reminds us to re-examine the role of the exponential function in architectural design.

\begin{figure}[h]
    \centering
    \includegraphics[width=5.5in]{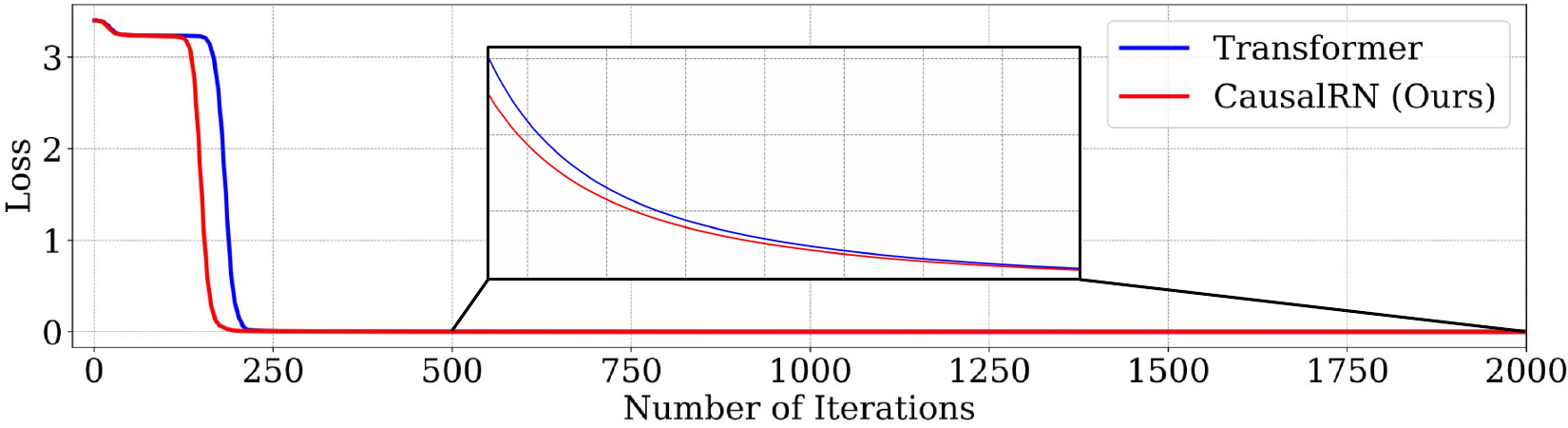}
    \caption{
    Comparison of learning curves.
    We zoom in from 500 to 2000 for closer observation.
    }
    \label{figure_comparison_2}
\end{figure}

\clearpage
\paragraph{Comparing Learning Curves}
In Figure~\ref{figure_comparison_2}, we set the string size to 128 and directly compare the learning curves of the CausalRN and Transformer.
We notice that the curves are very similar, even after converging to near zero loss values. This implies that CausalRNs might have some underlying connection with Transformers.
The initial plateau and near vertical descents from both models is a clear indication of the phase change phenomenon \cite{grokking2023nanda}.

\begin{figure}[h]
    \centering
    \includegraphics[width=5.5in]{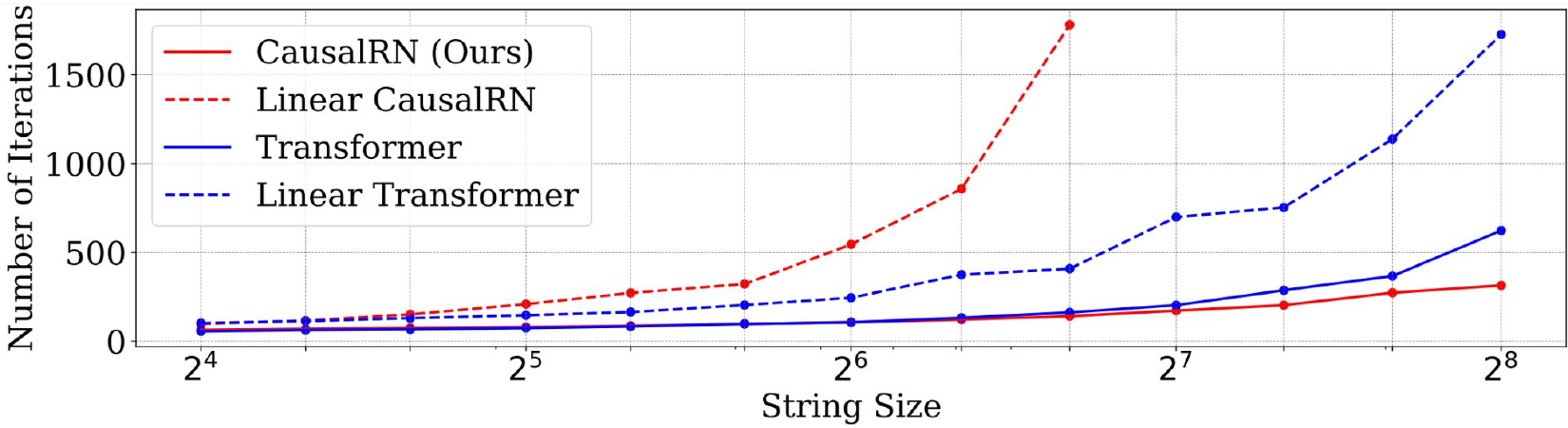}
    \caption{
    Convergence comparison.
    The y values correspond to the number of iterations until the model reaches 99\% accuracy.
    Linear CausalRN did not converge for string sizes \(\geq 2^7\).
    }
    \label{figure_comparison_1}
\end{figure}

\paragraph{Comparing Convergence}
\label{subsection_comparing_with_transformers}
In Figure~\ref{figure_comparison_1}, we vary the string sizes from \(2^4\) to \(2^8\) to observe the scaling property of four models.
Notably, simply by adding pre-activation normalization, CausalRNs change from hardly converging to converging faster than Transformers.
The linear models whose memory is reducible to fixed vectors constantly perform worse than their quadratic counterparts, aligned with prior observations \cite{copy2024jelassi}.
This result verifies our claim that applying pre-activation normalization recovers an irreducible memory pool, which in turn support effective in-context retrieval.

\section{Conclusion}
\label{section_conclusion}
In this paper, we introduced Causal Relation Network,
an all-MLP sequence modeling architecture that achieves Transformer-level copying capability.
Our findings highlight the importance of the exponential function and an irreducible memory pool for effective in-context retrieval.
Moreover, our experimental results suggest that Transformers' specific construct may not be the only way to excel at in-context retrieval.

\clearpage
\bibliography{references}

\clearpage
\appendix

\clearpage
\section{Numerically Stable Implementation for Linear Relation Networks}
\label{appendix_implement_LRN}
\begin{figure}[h]
    \centering
    \includegraphics[width=6in]{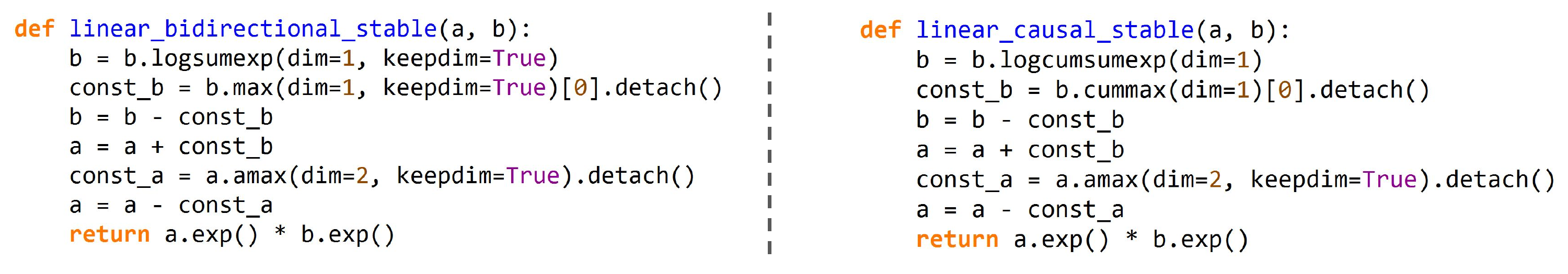}
    \caption{
    Code snippets for implementing numerically stable Linear BiRNs and Linear CausalRNs.
    }
    \label{figure_code_snippets}
\end{figure}

It is straightforward to implement Linear Bidirectional Relation Networks (BiRNs; see Appendix~\ref{appendix_birn}) and Linear Causal Relation Networks (CausalRNs) in PyTorch \cite{pytorch2019paszke}.
In Figure~\ref{figure_code_snippets}, we share code snippets for numerically stable implementations of both Linear BiRNs and Linear CausalRNs.
Variable \texttt{a} and \texttt{b} each has shape
\texttt{(batch\_size, num\_tokens, emb\_size)}.
Variable \texttt{a} corresponds to the set of vectors outputted by a linear layer,
and variable \texttt{b} corresponds to the set of vectors outputted by an affine layer.
The key insight is to use \texttt{logsumexp} or \texttt{logcumsumexp} to perform the reduction step before subtracting maximum values along certain axes.
This stability trick preserves collinearity. Therefore, as long as we apply post-reduction normalization, our stability trick is exact. The full code is at
\texttt{github.com/FrankCuiCN/causal-relation-networks}.

\clearpage
\section{Linear Causal Relation Networks}
\label{appendix_crn}
In this section, we train a Linear Causal Relation Network (CausalRN) to perform character-level language modeling.

We used WikiText-103, a text dataset derived from Wikipedia articles.
After character-level tokenization, the training set totals 522,243,436 tokens.
This allowed the models to converge within a single epoch.
We trained all models using a batch size of 320 for one epoch.
For optimization, we used Adam with \(1\times 10^{-4}\) learning rate.

\begin{figure}[h]
    \centering
    \includegraphics[width=6in]{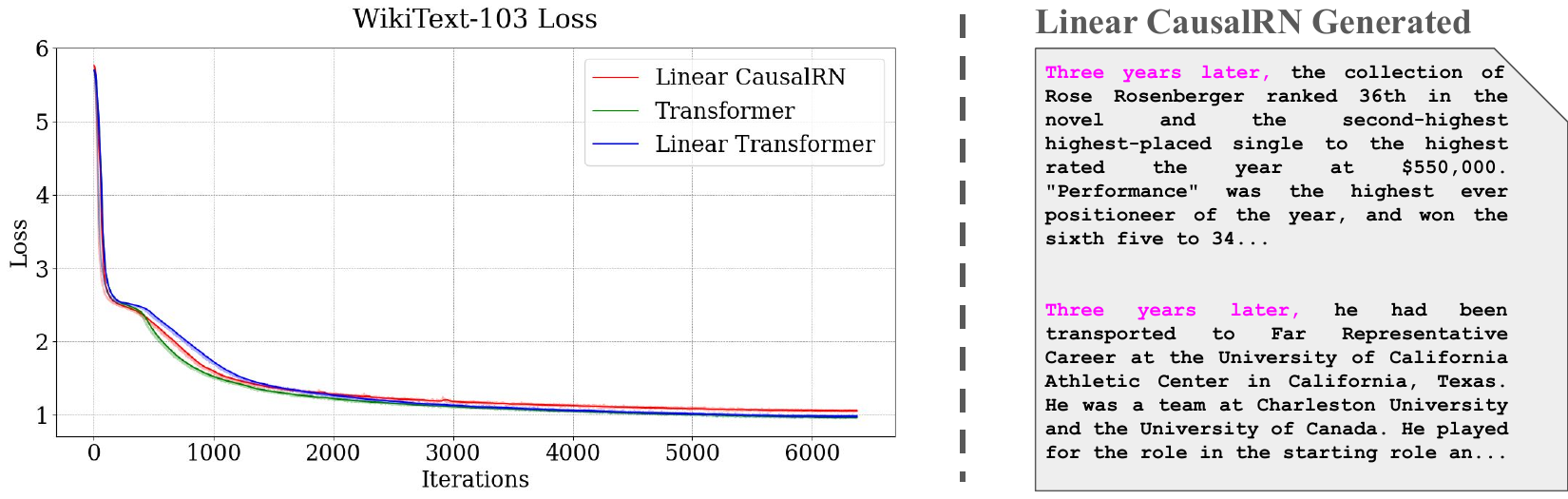}
    \caption{
    WikiText-103 results. Left: Loss over iterations. Right: Text samples generated by the Linear CausalRN model. The generation is conditioned on the phrase ``Three years later,''.
    }
    \label{wikitext}
\end{figure}

As shown in Figure~\ref{wikitext}, the Linear CausalRN converges similarly to Transformers\cite{transformers2017vaswani} and Linear Transformers in terms of cross-entropy loss.
This suggests that CausalRNs are valid models to perform autoregressive language modeling.

In Figure~\ref{wikitext}, we include two samples of text generated by the Linear CausalRN.
Considering this is a character-level model, this shows that the model is good at spelling and can capture some semantic dependencies.

\clearpage
\section{Linear Bidirectional Relation Networks}
\label{appendix_birn}
We use Linear Bidirectional Relation Networks (BiRNs) to implement a computer vision model.
BiRNs are bidirectional counterparts of CausalRNs.
Specifically, for vectors
$X = \{\textbf{x}_1, \textbf{x}_2, ..., \textbf{x}_n\}$
where $\mathbf{x}_i \in \mathbb{R}^{d_{e}},$ and
$f_\theta:\mathbb{R}^{2d_e} \rightarrow \mathbb{R}^{d_e},$
we define \textit{Bidirectional Relation Network (BiRN)} as
\begin{equation}
\label{eq_BiRN}
\mathbf{y}_{j} = \frac{1}{n} \sum_{i=1}^{n} f_\theta([\mathbf{x}_i ; \mathbf{x}_j]).
\end{equation}

The dataset we choose is CIFAR-5m\cite{cifar5m2021preetum}.
CIFAR-5m is an extended version of CIFAR-10 \cite{cifar102009krizhevsky}, including 5 million synthetic images. The abundance in data allows all models to converge within one epoch.
Following ViT \cite{vit2021alexey}, we arranged the images into 2 by 2 patches and added a \texttt{<CLS>} token to output the classification result.
We trained the model using a batch size of 320 for one epoch. For optimization, we used Adam with \(5 \times 10^{-4}\) learning rate.

\begin{figure}[h]
    \centering
    \includegraphics[width=5.5in]{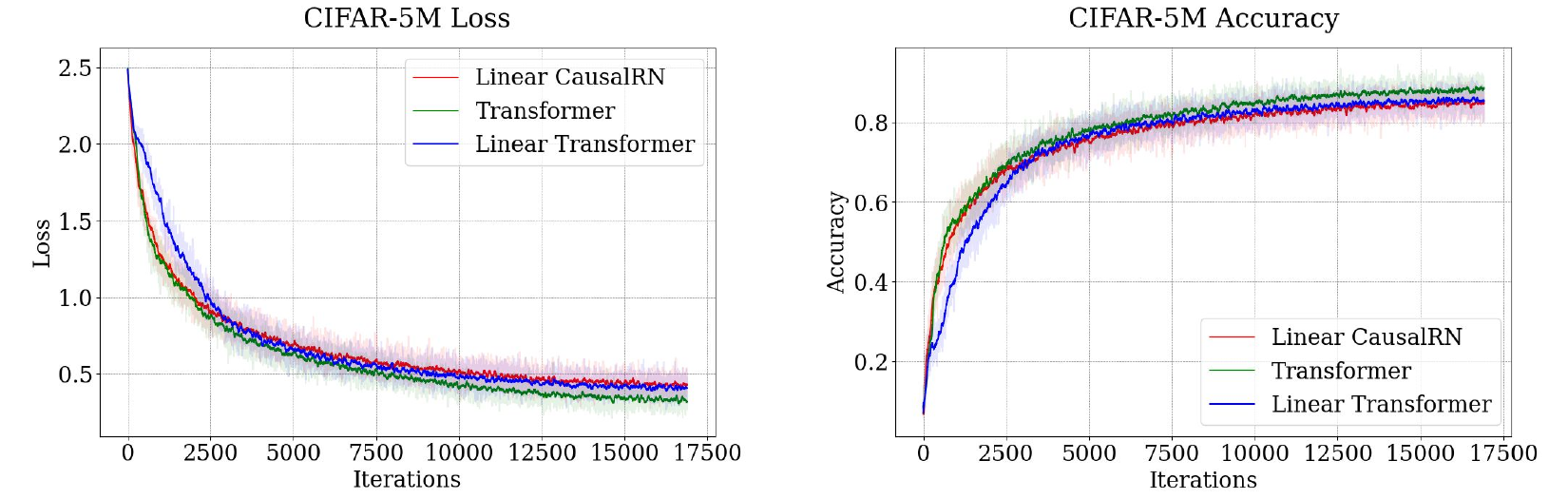}
    \caption{
    CIFAR-5M results. Left: Loss over iterations. Right: Accuracy over iterations.
    }
    \label{cifar5m}
\end{figure}

As shown in Figure~\ref{cifar5m}, the Linear BiRN converges similarly to Transformers\cite{transformers2017vaswani} and Linear Transformers in terms of cross entropy loss and accuracy.

\begin{figure}[h]
    \centering
    \includegraphics[width=5.5in]{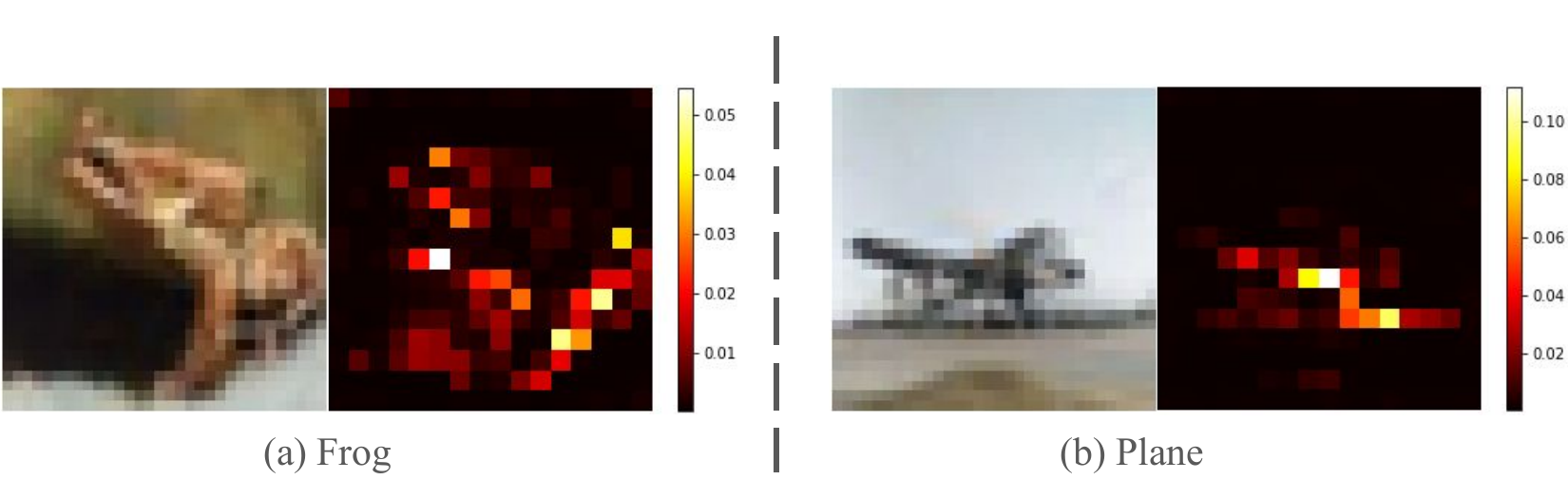}
    \caption{
    Interpretability result showing both the original images and the corresponding heat maps.
    }
    \label{interpretability}
\end{figure}

In Figure~\ref{interpretability}, we visualize heat maps extracted from a trained BiRN model.
The heat maps come from the 6\textsuperscript{th} layer and is showing how strongly each patch attends to the \texttt{<CLS>} token.
It shows that the model is able to focus on the main object and ignore background elements.

\clearpage
\section{Implementation Details}
\label{appendix_implementation_details}
\paragraph{Architecture}
We compared our CausalRN with Linear CausalRN, Transformers \cite{transformers2017vaswani}, and Linear Transformers\cite{lineartransf2020katharopoulos}.
To match the scale of the copying task, we set \(d_e=192\), which was \(150\%\) of 128 and \(75\%\) of 256. 
For both linear and quadratic CausalRNs, we chose \(d_h=d_e=192\).
The number of (Linear) CausalRN blocks is 12, totaling 1.34 million parameters.
For both linear and quadratic Transformers, we chose \(d_h=4 d_e=768\).
We used 12 (Linear) Transformer blocks, totaling 5.33 million parameters.
For both Transformers, a sweep for the attention heads were performed in \(\{1, 2, 4, 8\}\).
We found using a single head gave optimal performance under our choice of \(d_e=192\).

\paragraph{Initialization}
Following common practices, we initialized all biases to zero and all weights randomly following $\mathcal{N}(0, 0.02)$, except for output layers, which were further divided by the square root of the number of residual connections. Token embeddings and positional embeddings were trainable and randomly initialized following $\mathcal{N}(0, 1)$.

\paragraph{Optimization}
For optimization, we used Adam with \(\beta_1=0.9\) and \(\beta_2=0.999\).
We linearly warmed up the learning rate within the first 50 iterations.
Without data leakage, we performed a learning rate sweep in \(\{1 \times 10^{-5}, 5 \times 10^{-5}, 1 \times 10^{-4}, 5 \times 10^{-4}, 1 \times 10^{-3}\}\), and found \(5 \times 10^{-4}\) to be optimal for Transformers and \(1 \times 10^{-3}\) to be optimal for CausalRNs.
To ensure a meaningful comparison of convergence speed, we chose \(5 \times 10^{-4}\) since this was the minimum of both.
We note that this choice was optimal for Transformers and suboptimal for CausalRNs.

\paragraph{Training}
We trained all models on a single NVIDIA A100 GPU using a batch size of 320 for a maximum of 2000 iterations.
Each training run can be completed within one and a half hours.

\paragraph{Evaluation}
For each iteration step, we calculated the cross-entropy loss and average accuracy from 320 online samples to evaluate the models.
The accuracy was computed in parallel, not through autoregressive decoding.
We note that a 100\% parallel accuracy necessarily implies a 100\% autoregressive accuracy.
Both CausalRNs and Transformers eventually achieves 100\% parallel accuracy.

\end{document}